\documentclass[english,conference]{IEEEtran}
\usepackage[T1]{fontenc}
\usepackage[latin1]{inputenc}
\usepackage{float}
\usepackage{graphicx}

\makeatletter

\providecommand{\tabularnewline}{\\}
\floatstyle{ruled}
\newfloat{algorithm}{tbp}{loa}
\floatname{algorithm}{Algorithm}

 \usepackage{verbatim}

\usepackage{times}
\usepackage{algorithmic}

\usepackage{babel}
\makeatother
\begin{document}

\title{Localization of Simultaneous Moving Sound Sources for Mobile Robot
Using a Frequency-Domain Steered Beamformer Approach}

\author{\authorblockN{Jean-Marc Valin, Fran\c{c}ois Michaud, Brahim Hadjou,
Jean Rouat}\authorblockA{LABORIUS, Department of Electrical Engineering and Computer Engineering\\
Universit\'e de Sherbrooke, Sherbrooke (Quebec) CANADA, J1K 2R1\\
\{Jean-Marc.Valin, Francois.Michaud, Brahim.Hadjou, Jean.Rouat\}@USherbrooke.ca}
}

\maketitle
\begin{abstract}
\footnotetext{\copyright 2004 IEEE.  Personal use of this material is permitted. Permission from IEEE must be obtained for all other uses, in any current or future media, including reprinting/republishing this material for advertising or promotional purposes, creating new collective works, for resale or redistribution to servers or lists, or reuse of any copyrighted component of this work in other works.}
Mobile robots in real-life settings would benefit from being able
to localize sound sources. Such a capability can nicely complement
vision to help localize a person or an interesting event in the environment,
and also to provide enhanced processing for other capabilities such
as speech recognition. In this paper we present a robust sound source
localization method in three-dimensional space using an array of 8
microphones. The method is based on a frequency-domain implementation
of a steered beamformer along with a probabilistic post-processor.
Results show that a mobile robot can localize in real time multiple
moving sources of different types over a range of 5 meters with a
response time of 200 ms.
\end{abstract}

\section{Introduction}

The sense of hearing is quite important in providing information in
a real life environment: it can draw attention to particular and discriminate
events in the world that can be further analyzed using other senses
such as vision, or it allows to exchange information through language.
For those who do not have hearing impairments, it is hard to imagine
going a day without being able to hear, especially given the fact
that we are moving in many different environments (indoor and outdoor).

Signal processing research that address artificial audition is often
geared toward specific tasks such as speaker tracking for videoconferencing.
However, artificial hearing for mobile robots is still in its infancy.
The SAIL robot uses one microphone to develop online audio-driven
behaviors \cite{zhang-weng2001}. The robot ROBITA uses two microphones
to follow a conversation between two people \cite{matsusaka-tojo-kubota-furukawa-tamiya-hayata-nakano-kobayashi99}.
SIG, a humanoid robot uses two pairs of microphones; one pair is installed
on both sides of the head, while the other pair is placed inside the
head to record internal sounds (such as motor noise) for noise cancellation
\cite{nakadai-okuno-kitano2002,okuno-nakadai-kitano2002}. Like humans,
these last two robots use binaural localization, i.e. the ability
to locate the source of sound in three dimensional space.

It is difficult to localize sounds with only two input sources. The
human auditory system accounts for the acoustic shadow of the head
and the ridges of the outer ear. Without this ability, only localization
in two dimensions is possible without the possibility to distinguish
if the sounds come from the front or the back. Also, it may be difficult
to obtain high-precision readings when the sound source is in the
same axis of the pair of microphones.

Robots are not inherently limited to two microphones; we decided to
use more microphones to better approach the localization abilities
of the human auditory system. This way, increased resolution can be
obtained in three-dimensional space. This also means increased robustness,
since multiple signals greatly helps reduce the effects of noise (instead
of trying to isolate the noise source by putting sensors inside the
robot's head, as with SIG) and discriminate multiple sound sources.
There are already robots available with more than two microphones;
the Sony SDR-4X has seven.

An artificial audition system can be used for three things: 1) localizing
sound sources, 2) separating sound sources in order to process only
signals that are relevant to a particular event in the environment,
and 3) processing sound sources to extract useful information from
the environment (like speech recognition for instance). This paper
focuses on sound source localization. In previous work \cite{ValinIROS2003},
we presented a method based on time delay of arrival (TDOA) estimation.
The method works for far-field and near-field sound sources and was
validated using a Pioneer 2 mobile robotic platform. 

In this paper, we present an approach with the same objective, but
is based on a frequency-domain beamformer that is steered in all possible
directions to detect sources. Instead of measuring TDOAs and then
converting to a position, the search is performed in a single step.
This makes the system more robust, especially in the case where an
obstacle prevents one or more microphones from properly receiving
the signals. The results are then enhanced by probability-based post-processing
which prevents false detection of sources. This makes the system sensitive
enough for simultaneous localization of multiple moving sound sources. 

The paper is organized as follows. Section \ref{sec:System-overview}
presents a brief overview of the system and Section \ref{sec:Localization-by-steered}
describes our frequency-domain implementation of a steered beamformer.
Section \ref{sec:Probabilistic-post-processing} explains how we enhance
the results from the beamformer using a probabilistic post-processor,
followed by experimental results in Section \ref{sec:Results}.

\section{System overview}

\label{sec:System-overview}
\begin{figure}[th]
\center{\includegraphics[%
  width=0.70\columnwidth,
  keepaspectratio]{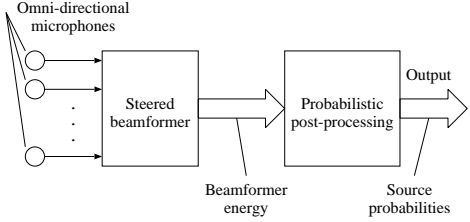}}

\caption{Overview of the system\label{cap:Overview-of-the-system}}
\end{figure}

The proposed localization system as shown
in Figure \ref{cap:Overview-of-the-system} is composed of three parts:

\begin{itemize}
\item A microphone array;
\item A memoryless localization algorithm based on a steered beamformer;
\item A probability-based post-processor.
\end{itemize}

The microphone array is composed of a number of omni-directional elements
mounted on the robot. The signals are used by a beamformer that is
steered in all possible directions in order to maximize the output
power. The initial localization performed by the steered beamformer
is then used as the input of a post-processing stage that uses Bayesian
probability rules to compute the probability of source presence for
every directions.

The output of the localization can be used to direct the robot attention
to the source. It can also be used as part of a source separation
algorithm to isolate the sound coming from a single source \cite{ValinICASSP2004}.

\section{Localization by steered beamformer}

\label{sec:Localization-by-steered}The basic idea behind the steered
beamformer approach to source localization is to direct a beamformer
in all possible directions and look for maximal output. For this task,
we try to maximize the output power of a simple delay-and-sum beamformer.

\subsection{Delay-and-sum beamformer}

The output of an $M$-microphone delay-and-sum beamformer is defined
as:

\begin{equation}
y(n)=\sum_{m=0}^{M-1}x_{m}\left(n-\tau_{m}\right)\label{eq:delay-and-sum}\end{equation}
where $x_{m}\left(n\right)$ is the signal from the $m^{th}$ microphone
and $\tau_{m}$ is the delay of arrival for that microphone. The output
energy of the beamformer over a frame of length $L$ is thus given
by:\begin{eqnarray}
E & = & \sum_{n=0}^{L-1}\left[y(n)\right]^{2}\nonumber \\
 & = & \sum_{n=0}^{L-1}\left[x_{0}\left(n-\tau_{0}\right)+\ldots+x_{M-1}\left(n-\tau_{M-1}\right)\right]^{2}\label{eq:beamformer-energy}\end{eqnarray}

Assuming that one sound source is present, we can see that $E$ will
be maximal when the delays $\tau_{m}$ are such that the microphone
signals are in phase (and therefore add constructively).

There is, however, a problem with that technique in that energy peaks
are very wide \cite{Duraiswami2001}, which means that the resolution
is poor. Moreover, in the case of multiple sources, it makes it more
likely to have sources responses overlap. 

One way to narrow the peaks is to whiten the microphone signals prior
to computing the energy \cite{Omologo}. Unfortunately, the coarse-fine
search methods as proposed in \cite{Duraiswami2001} cannot be used
because the narrow peaks can be missed during the coarse search. Therefore,
a fine search is necessary, which requires increased computing power.
It is however possible to reduce the amount of computation by calculating
the beamformer energy in the frequency domain. This also has the advantage
of making the whitening of the signal easier.

We first notice that the beamformer output energy in Equation \ref{eq:beamformer-energy}
can be expanded as:

\begin{eqnarray}
E & = & \sum_{m=0}^{M-1}\sum_{n=0}^{L-1}x_{m}^{2}\left(n-\tau_{m}\right)\nonumber \\
 & + & 2\sum_{m_{1}=0}^{M-1}\sum_{m_{2}=0}^{m_{1}-1}\sum_{n=0}^{L-1}x_{m_{1}}\left(n-\tau_{m_{1}}\right)x_{m_{2}}\left(n-\tau_{m_{2}}\right)\label{eq:beamformer-energy-expand}\end{eqnarray}
which in turn can be rewritten in terms of cross-correlations:

\begin{equation}
E=K+2\sum_{m_{1}=0}^{M-1}\sum_{m_{2}=0}^{m_{1}-1}R_{x_{m_{1}},x_{m_{2}}}\left(\tau_{m_{1}}-\tau_{m_{2}}\right)\label{eq:energy-xcorr}\end{equation}
where $K=\sum_{m=0}^{M-1}\sum_{n=0}^{L-1}x_{m}^{2}\left(n-\tau_{m}\right)$
is nearly constant with respect to the $\tau_{m}$ delays and can
thus be ignored when maximizing $E$. The cross-correlation function
can be approximated in the frequency domain as:

\begin{equation}
R_{ij}(\tau)\approx\sum_{k=0}^{L-1}X_{i}(k)X_{j}(k)^{*}e^{\jmath2\pi k\tau/L}\label{eq:TDOA_correlation_freq}\end{equation}
where $X_{i}(k)$ is the discrete Fourier transform of $x_{i}[n]$,
$X_{i}(k)X_{j}(k)^{*}$ is the cross-spectrum of $x_{i}[n]$ and $x_{j}[n]$
and $(\cdot)^{*}$ denotes the complex conjugate. The power spectra
and cross-power spectra are computed on overlapping windows (50\%
overlap) of $L=1024$ samples at 48 kHz. Once the $R_{ij}(\tau)$
are precomputed, it is possible to compute $E$ using only $N(N-1)/2$
lookup and accumulation operations.

Because of the reduced complexity, it is possible to use two different
source detectors; a short-term one for percussive noise like dropped
objects or handclaps and a medium-term one for speech and other continuous
sounds.

For each estimator, the $R_{ij}(\tau)$ are computed by averaging
the cross-power spectra $X_{i}(k)X_{j}(k)^{*}$ over two different
time periods. In our implementation, the short- and medium-term estimators
average the cross-power spectra over 4 frames (40 ms) and 20 frames
(200 ms), respectively. The averaging also means that the overall
system has a response time of 200 ms.

\subsection{Spectral weighting\label{sub:Spectral-weighting}}

As stated in the previous subsection, we chose to whiten the signal
prior to computing the beamformer energy. In the frequency domain,
the whitened cross-correlation is thus computed as:\begin{equation}
R_{ij}^{(w)}(\tau)\approx\sum_{k=0}^{L-1}\frac{X_{i}(k)X_{j}(k)^{*}}{\left|X_{i}(k)\right|\left|X_{j}(k)\right|}e^{\jmath2\pi k\tau/L}\label{eq:TDOA_correlation_whitened}\end{equation}

While it produces much sharper cross-correlation peaks, the whitened
cross-correlation has a drawback. Each frequency bin of the spectrum
contributes the same amount to the final correlation, even if the
signal at that frequency is dominated by noise. This makes the system
less robust to noise, while making detection of voice (which has a
narrow bandwidth) more difficult.

In order to resolve the problem, we developed a weighting function
for the spectrum. This function gives more weight to regions in the
spectrum where the local signal-to-noise ratio (SNR) is the highest.
Let $Y(k)$ be the mean power spectral density for all the microphones
at a given time and $Y_{N}(k)$ be a noise estimate based on the time
average of previous $Y(k)$. We define a noise masking weight by:\begin{equation}
w(k)=\left\{ \begin{array}{ll}
1 & ,\: Y(k)\leq Y_{N}(k)\\
\left(\frac{Y(k)}{Y_{N}(k)}\right)^{\gamma} & ,\: Y(k)>Y_{N}(k)\end{array}\right.\label{eq:noise_weighting2}\end{equation}
where the exponent $0<\gamma<1$ gives more weight to regions where
the signal is much higher than the noise. For our system, we empirically
set $\gamma$ to $0.1$. The resulting enhanced cross-correlation
is defined as:\begin{equation}
R_{ij}^{(e)}(\tau)=\sum_{k=0}^{L-1}\frac{w^{2}(k)X_{i}(k)X_{j}(k)^{*}}{\left|X_{i}(k)\right|\left|X_{j}(k)\right|}e^{\jmath2\pi k\tau/L}\label{eq:TDOA_correlation_weighted}\end{equation}

\subsection{Direction search on spherical grid}

\begin{figure}[th]
\center{\includegraphics[%
  width=0.29\columnwidth,
  height=0.29\columnwidth]{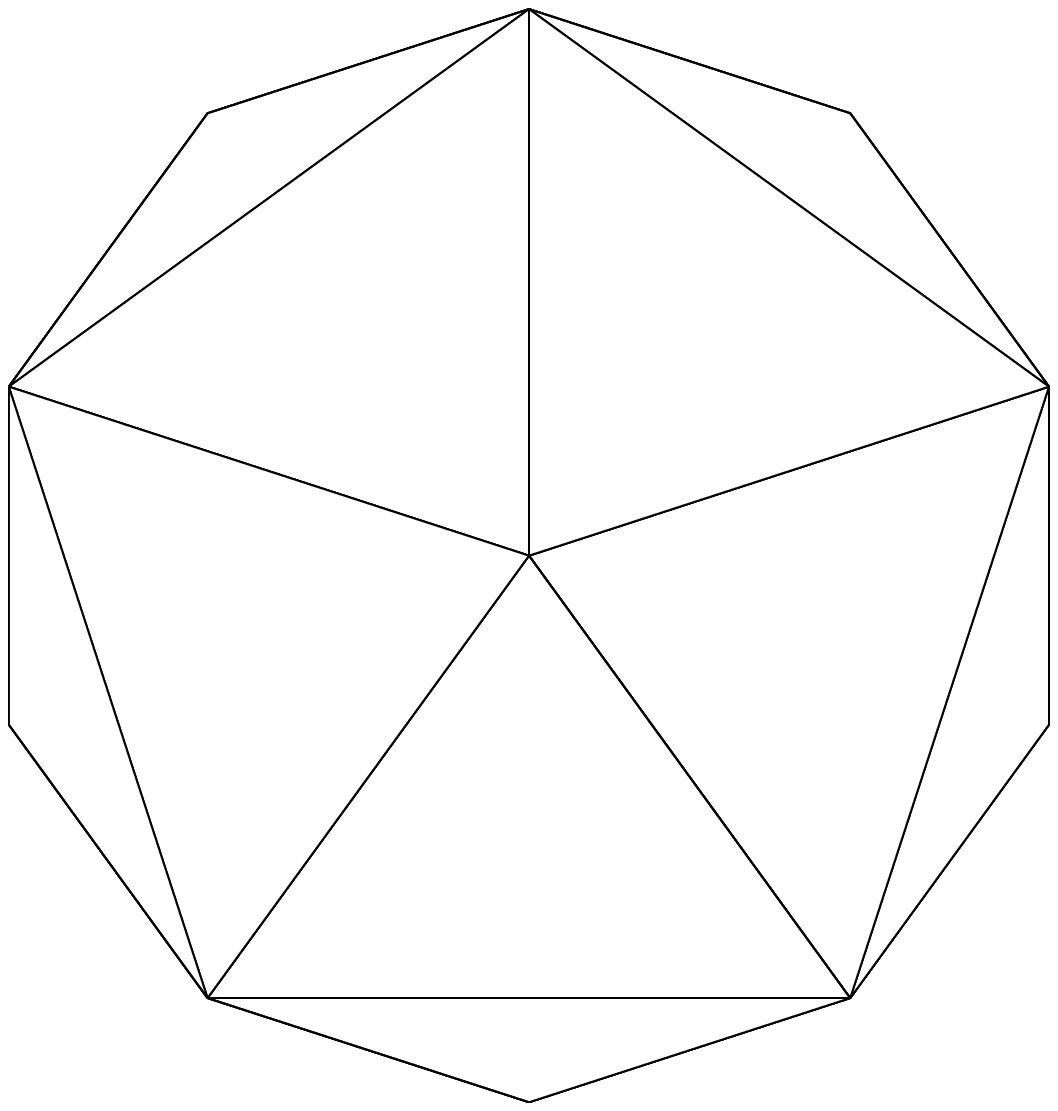}\hspace{3mm}\includegraphics[%
  width=0.29\columnwidth,
  height=0.29\columnwidth]{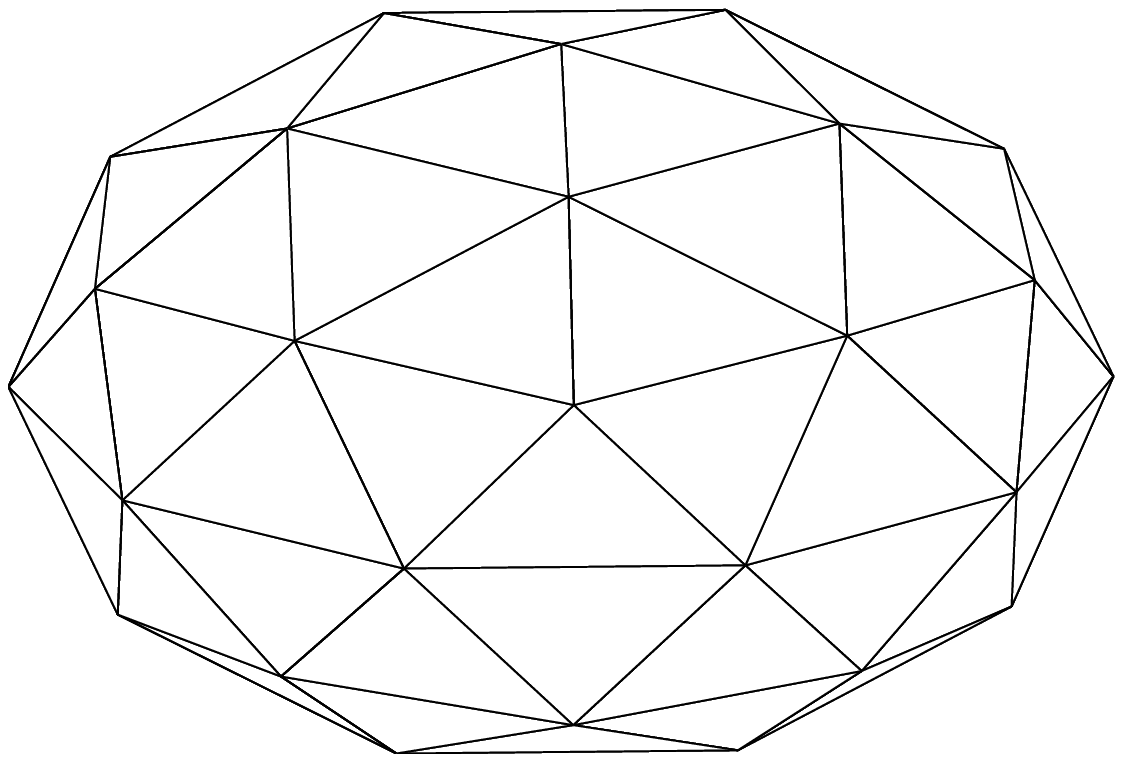}\includegraphics[%
  width=0.32\columnwidth,
  height=0.32\columnwidth]{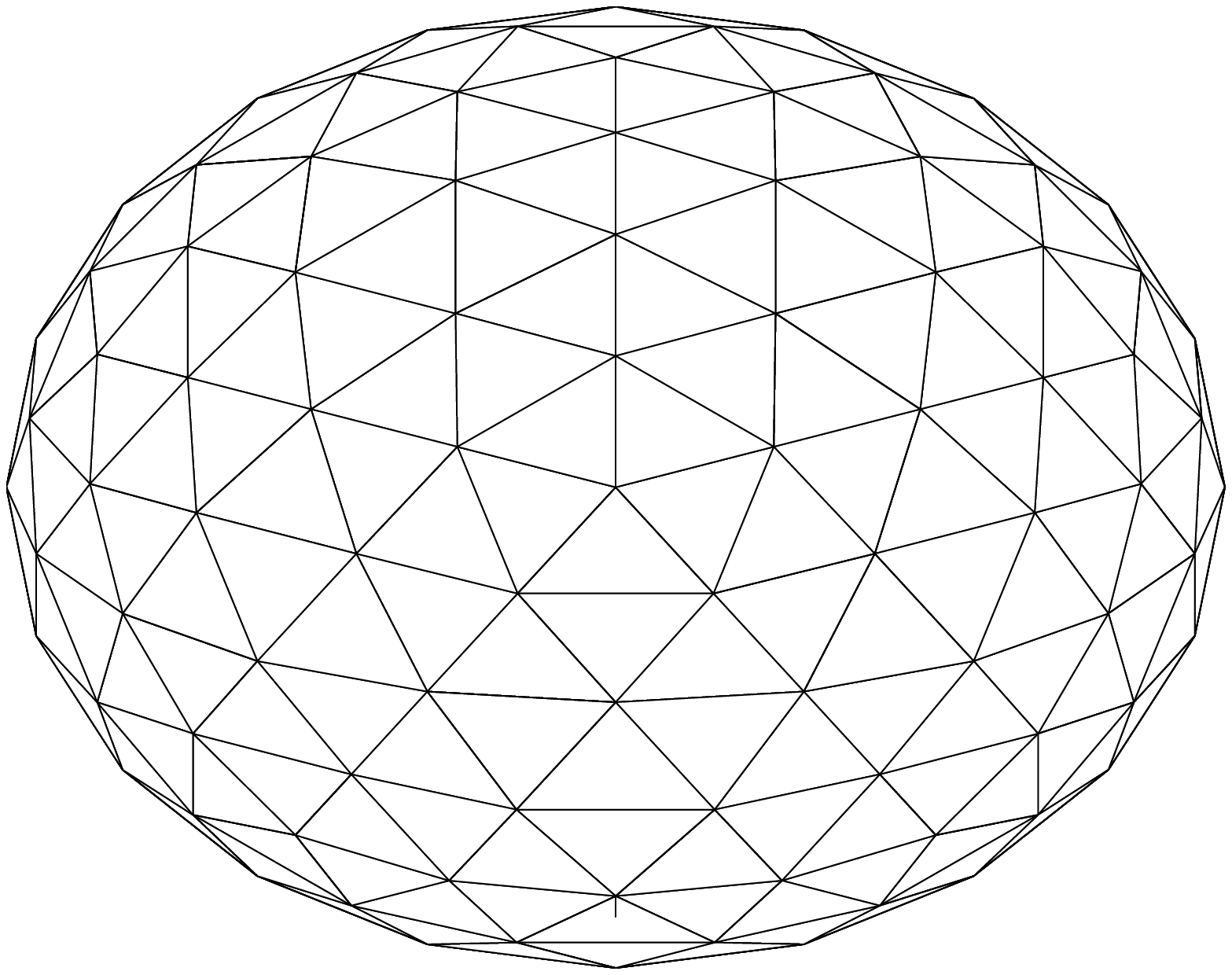}}

\caption{Recursive subdivision (2 levels) of a triangular element\label{cap:Recursive-subdivision}}
\end{figure}

In order to reduce the computation required and to make the system
isotropic, we define a uniform triangular grid for the surface of
a sphere. In order to create the grid, we start from an initial icosahedral
grid \cite{Giraldo97}. Each triangle in the initial 20-element grid
is then recursively subdivided into four smaller triangles as shown
in Figure \ref{cap:Recursive-subdivision}. The resulting grid is
composed of 5120 triangles and 2562 points. The beamformer energy
is then computed for the hexagonal region associated with each of
these points. Each of the 2562 regions covers a radius of about $2.5^{\circ}$
around its center, which means that it introduces at most an error
of $2.5^{\circ}$.

Once the cross-correlations $R_{ij}^{(e)}(\tau)$ are computed, the
search for the best direction on the grid is performed as described
by Algorithm \ref{cap:Steered-beamformer-direction}.

\begin{algorithm}[th]
\begin{algorithmic}

\FORALL{grid index $d$}

\STATE $E_d \leftarrow 0$

\FORALL{microphone pair $ij$}

\STATE $\tau \leftarrow lookup(d,ij)$

\STATE $E_d \leftarrow E_d + R^{(e)}_{ij}(\tau)$

\ENDFOR

\ENDFOR

\STATE \textit{direction of source} $\leftarrow \textrm{argmax}_d\ (E_d)$

\end{algorithmic}

\caption{Steered beamformer direction search\label{cap:Steered-beamformer-direction}}
\end{algorithm}
In Algorithm \ref{cap:Steered-beamformer-direction}, \emph{lookup}
is a precomputed table of the time delay of arrival (TDOA) for each
microphone pair and each direction on the sphere. By making the far-field
assumption \cite{ValinIROS2003}, the TDOA in samples is computed
as:\begin{equation}
\tau_{ij}=\frac{F_{s}}{c}\left(\vec{\mathrm{\mathbf{x}}}_{i}-\mathrm{\vec{\mathrm{\mathbf{x}}}}_{j}\right)\cdot\vec{\mathbf{u}}\label{eq:TDOA-far-field}\end{equation}
where $\vec{\mathrm{\mathbf{x}}}_{i}$ is the position of microphone
$i$, $\vec{\mathbf{u}}$ is a unit-vector that points in the direction
of the source, $c$ is the speed of sound and $F_{s}$ is the sampling
rate.

For an array of $M$ microphones and an $N$-element grid, the algorithm
requires $M(M-1)N$ table memory accesses and $M(M-1)N/2$ additions.
In the proposed configuration ($N=2562$, $M=8$), the accessed data
can be made to fit entirely in a modern processor's L2 cache.

The algorithm described above is able to find the loudest source present
by maximizing the energy of a steered beamformer. In order to localize
other sources that may be present, we remove the contribution of the
first source to the cross-correlations. The process is then repeated,
which leads to Algorithm \ref{cap:Localization-of-multiple-sources}.

\begin{algorithm}[th]
\begin{algorithmic}

\FOR{$k=1$ to desired number of sources}

\STATE $D_k \leftarrow \textrm{Steered beamformer direction search}$

\FORALL{microphone pair $ij$}

\STATE $\tau \leftarrow lookup(D_k,ij)$

\STATE $R^{(e)}_{ij}(\tau) = 0$

\ENDFOR

\ENDFOR

\end{algorithmic}

\caption{Localization of multiple sources\label{cap:Localization-of-multiple-sources}}
\end{algorithm}

The number of desired sources is a constant for each estimator. We
consider that the short-term estimator is able to locate at most two
sources at the same time, while the medium-term estimator is able
to detect four sources. When less sources are present this leads to
false detection of a source. That problem is handled by the probabilistic
post-processing described next.

\section{Probabilistic post-processing}

\label{sec:Probabilistic-post-processing}In order to prevent false
detection of sources and keep the system sensitive enough to weak
sources, we introduce a post-processing step that provides some smoothing
in time, while combining the results of the short- and medium-term
estimators. Using the same quantized sphere as in the previous section,
we associate a probability of source presence to each region of the
grid (we omit the grid region index for clarity). We note $H_{1}^{n}$
the hypothesis of source presence at discrete time $n$ and $H_{0}^{n}$
the hypothesis of no source being present at that time. Also, the
steered beamformer observation for time $n$ is denoted $o_{n}$,
with $\mathbf{O}_{n}=\left(o_{1},o_{2},\ldots,o_{n}\right)$ the set
of all observations up to time $n$.

We first introduce an instantaneous probability estimation that uses
the results of the steered beamformer of Section \ref{sec:Localization-by-steered}.
The idea is that the higher the output energy of the beamformer, the
more likely that a source is present. We thus approximate the instantaneous
probability of a source being present as:\begin{equation}
P\left(H_{1}^{n}\left|o_{n}\right.\right)=\max\left[1-\exp\left(1-\frac{E}{E_{min}}\right),p_{min}\right]\label{eq:instantaneous-prob}\end{equation}
where $E$ is the energy at the output of the beamformer, $E_{min}$
is an energy threshold corresponding to the value when no source is
present, and $p_{min}$ is the minimal probability we want to assign
for a source that is detected by the steered beamformer (with $p_{min}=0.1$).
In the case where there is no source detected by the beamformer at
a certain point, we assign a floor probability $p_{floor}=0.005$
that accounts for the possibility that the beamformer does not detect
anything even though a sound source is present.

\subsection{Temporal integration\label{sub:Temporal-integration}}

At time $N$, we use Bayes' rule to express the probability of source
presence given all observations as:

\begin{equation}
P\left(H_{1}^{n}\left|\mathbf{O}_{n}\right.\right)=\frac{P\left(\left.\mathbf{O}_{n}\right|H_{1}^{n}\right)P\left(H_{1}^{n}\right)}{P\left(\mathbf{O}_{n}\right)}\label{eq:prob_bayes}\end{equation}

Because the energy of the steered beamformer is computed on non-overlapping
segments, we assume conditional independence of the observations with
respect to the presence or absence of a source. We can thus rewrite
Equation \ref{eq:prob_bayes} as:

\begin{eqnarray}
P\left(H_{1}^{N}\left|\mathbf{O}_{n}\right.\right) & = & \frac{P\left(\left.\mathbf{O}_{n-1}\right|H_{1}^{n}\right)P\left(\left.o_{n}\right|H_{1}^{n}\right)p_{1}}{P\left(\mathbf{O}_{n}\right)}\nonumber \\
 & = & \frac{P\left(H_{1}^{n}\left|\mathbf{O}_{n-1}\right.\right)P\left(H_{1}^{n}\left|o_{n}\right.\right)}{p_{1}}\nonumber \\
 & \cdot & \frac{P\left(\mathbf{O}_{n-1}\right)P\left(o_{n}\right)}{P\left(\mathbf{O}_{n}\right)}\label{eq:prob-bayes-separ}\end{eqnarray}

where $p_{1}=P\left(H_{1}^{n}\right)=P\left(H_{1}\right)$ is the
constant \emph{a priori} probability of source presence. Similarly,
it follows that the complementary probability is given by:

\begin{eqnarray}
P\left(\! H_{0}^{n}\left|\mathbf{O}_{N}\right.\!\right) & = & \frac{\left[\!1-P\left(\! H_{1}^{n}\left|\mathbf{O}_{n-1}\right.\!\right)\right]\left[\!1-P\left(\! H_{1}^{n}\left|o_{n}\right.\!\right)\right]}{\left(1-p_{1}\right)}\nonumber \\
 & \cdot & \frac{P\left(\mathbf{O}_{n-1}\right)P\left(o_{n}\right)}{P\left(\mathbf{O}_{n}\right)}\label{eq:prob-bayes-separ-H0}\end{eqnarray}

We assume that the transitions between $H_{0}$ and $H_{1}$ can be
modeled as a first order Markov process with transition probabilities
$\alpha_{ij}=P\left(\left.H_{j}^{n}\right|H_{i}^{n-1}\right)$. This
leads to:

\begin{eqnarray}
P\left(H_{1}^{n}\left|\mathbf{O}_{n-1}\right.\right) & = & \alpha_{01}\left[1-P\left(H_{1}^{n-1}\left|\mathbf{O}_{n-1}\right.\right)\right]\nonumber \\
 & + & \alpha_{11}P\left(H_{1}^{n-1}\left|\mathbf{O}_{n-1}\right.\right)\label{eq:markov-prob-H1}\end{eqnarray}
For this work, we use $\alpha_{01}=0.00004,\:\alpha_{11}=0.992$ for
the short-term estimator and $\alpha_{01}=0.0002,\:\alpha_{11}=0.96$
for the medium-term estimator. The reason for the differences in values
is that the medium-term estimator is updated less often.

In order to avoid computing $P\left(\mathbf{O}_{n}\right)$, $P\left(\mathbf{O}_{n-1}\right)$
and $P\left(o_{n}\right)$ terms that do not depend on $H_{0}$ or
$H_{1}$, we introduce the unnormalized probabilities $\pi\left(H_{1}^{N}\left|\mathbf{O}_{n}\right.\right)$
and $\pi\left(H_{0}^{N}\left|\mathbf{O}_{n}\right.\right)$ that omit
these terms. For example, from Equation \ref{eq:prob-bayes-separ},
we have the unnormalized probability:\begin{equation}
\pi\left(H_{1}^{N}\left|\mathbf{O}_{n}\right.\right)=\frac{P\left(H_{1}^{n}\left|\mathbf{O}_{n-1}\right.\right)P\left(H_{1}^{n}\left|o_{n}\right.\right)}{p_{1}}\label{eq:prob-bayes-separ-unnorm}\end{equation}

From there, it is easy to compute $P\left(H_{1}^{N}\left|\mathbf{O}_{n}\right.\right)$
as:\begin{eqnarray}
P\left(H_{1}^{N}\left|\mathbf{O}_{N}\right.\right) & = & \frac{\pi\left(H_{1}^{N}\left|\mathbf{O}_{N}\right.\right)}{\pi\left(H_{1}^{N}\left|\mathbf{O}_{N}\right.\right)+\pi\left(H_{0}^{N}\left|\mathbf{O}_{N}\right.\right)}\nonumber \\
 & = & \frac{1}{1+\frac{\pi\left(H_{0}^{N}\left|\mathbf{O}_{N}\right.\right)}{\pi\left(H_{1}^{N}\left|\mathbf{O}_{N}\right.\right)}}\label{eq:prob-normalize}\end{eqnarray}

\subsection{Combination of estimator probabilities}

After using the temporal integration method to derive the short-term
and medium-term estimators, the last step consists of combining these
probabilities to infer a unique probability of source presence. Let
$\mathbf{O}^{s},\mathbf{O}^{m}$ respectively be all the observations
made by the short- and medium-term estimators up to a certain time,
we can first write using Bayes' rule:\begin{equation}
P\left(\left.H_{1}\right|\mathbf{O}^{s},\mathbf{O}^{m}\right)=\frac{P\left(\mathbf{O}^{s},\mathbf{O}^{m}\left|H_{1}\right.\right)P\left(H_{1}\right)}{P\left(\mathbf{O}^{s},\mathbf{O}^{m}\right)}\label{eq:prob-fusion}\end{equation}
Unfortunately, we cannot assume that $\mathbf{O}^{s}$ and $\mathbf{O}^{m}$
are conditionally independent. To represent that, we approximate the
combined probability as a weighted geometric average of two hypotheses:
1) complete independence of $\mathbf{O}^{s}$ and $\mathbf{O}^{m}$
and 2) equivalence of $\mathbf{O}^{s}$ and $\mathbf{O}^{m}$.

If we consider the hypothesis of complete conditional independence
of the different estimators, we have:\begin{eqnarray}
P_{i}\left(\!\left.H_{1}\right|\mathbf{O}^{s},\mathbf{O}^{m}\!\right) & = & \frac{P\left(\!\mathbf{O}^{s}\left|H_{1}\right.\!\right)P\left(\!\mathbf{O}^{m}\left|H_{1}\right.\!\right)p_{1}}{P\left(\mathbf{O}^{s},\mathbf{O}^{m}\right)}\nonumber \\
 & = & \frac{P\left(\!\left.H_{1}\right|\mathbf{O}^{s}\!\right)P\left(\!\left.H_{1}\right|\mathbf{O}^{m}\!\right)}{p_{1}}\nonumber \\
 & \cdot & \frac{P\left(\mathbf{O}^{s}\right)P\left(\mathbf{O}^{m}\right)}{P\left(\mathbf{O}^{s},\mathbf{O}^{m}\right)}\label{eq:prob-fusion-indep}\end{eqnarray}
The complementary probability $P_{i}\left(\!\left.H_{0}\right|\mathbf{O}^{s},\mathbf{O}^{m}\!\right)$
can be estimated similarly.

In addition to the complete conditional independence hypothesis, we
consider the case where $\mathbf{O}^{s},\mathbf{O}^{m}$ bring exactly
the same information about source presence or absence. In that case,
all probabilities should be equal, so we rewrite the probability as:\begin{equation}
P_{d}\left(\!\left.H_{1}\right|\!\mathbf{O}^{s},\mathbf{O}^{m}\right)=\sqrt{P\left(\!\left.H_{1}\right|\!\mathbf{O}^{s}\right)P\left(\!\left.H_{1}\right|\!\mathbf{O}^{m}\right)}\label{eq:prob-fusion-depend}\end{equation}

The reality lies in between the situation described by Equations \ref{eq:prob-fusion-indep}
and \ref{eq:prob-fusion-depend}. We express the combined probability
estimation as:\begin{eqnarray}
P\left(\left.H_{1}\right|\mathbf{O}^{s},\mathbf{O}^{m}\right) & \approx & \left[P_{d}\left(\left.H_{1}\right|\mathbf{O}^{s},\mathbf{O}^{m}\right)\right]^{\beta}\nonumber \\
 & \cdot & \left[P_{i}\left(\left.H_{1}\right|\mathbf{O}^{s},\mathbf{O}^{m}\right)\right]^{1-\beta}\label{eq:fusion-geom-mean}\end{eqnarray}
where $0\leq\beta\leq1$ expresses the degree of dependence between
the observations ($\beta=0$ is complete independence), $P_{i}$ is
the probability assuming complete independence and $P_{d}$ is the
probability assuming equivalence of $\mathbf{O}^{s}$ and $\mathbf{O}^{m}$.
For this paper, we use $\beta=0.7$.

Using the same unnormalized probabilities defined in the Section \ref{sub:Temporal-integration},
we have: 

\begin{equation}
P\left(\left.H_{1}\right|\mathbf{O}^{s},\mathbf{O}^{m}\right)\approx\frac{1}{1+\frac{\pi\left(\left.H_{0}\right|\mathbf{O}^{s},\mathbf{O}^{m}\right)}{\pi\left(\left.H_{1}\right|\mathbf{O}^{s},\mathbf{O}^{m}\right)}}\label{eq:fusion-unnorm}\end{equation}
where:\begin{eqnarray}
\!\!\!\!\pi\!\left(\!\left.H_{1}\right|\!\mathbf{O}^{s}\!,\!\mathbf{O}^{m}\!\right) & = & \frac{\left[P\!\left(\!\left.H_{1}\!\right|\!\mathbf{O}^{s}\!\right)\! P\!\left(\!\left.H_{1}\!\right|\!\mathbf{O}^{m}\!\right)\right]^{1-\frac{\beta}{2}}}{p_{1}^{1-\beta}}\label{eq:fusion-unnorm-def1}\\
\!\!\!\!\pi\!\left(\!\left.H_{0}\right|\!\mathbf{O}^{s}\!,\!\mathbf{O}^{m}\!\right) & = & \frac{\left[\!\left(1\!\!-\!\! P\!\left(\!\left.H_{1}\!\right|\!\mathbf{O}^{s}\!\right)\!\right)\!\left(1\!\!-\!\! P\!\left(\!\left.H_{1}\!\right|\!\mathbf{O}^{m}\!\right)\!\right)\!\right]^{1\!-\!\frac{\beta}{2}}}{\left(1-p_{1}\right)^{1-\beta}}\label{eq:fusion-unnorm-def2}\end{eqnarray}

Our choice of the geometric mean is based on the fact that the probabilities
can have a very wide dynamic range that is not suitable for the arithmetic
mean.

\section{Results}

\label{sec:Results}
\begin{figure}[th]
\center{\includegraphics[%
  width=0.90\columnwidth]{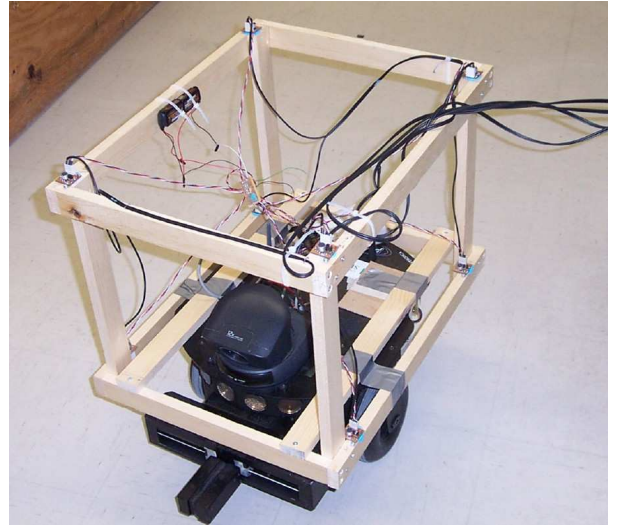}}

\caption{Pioneer 2 robot with an array of eight microphones\label{cap:array-pioneer2}}
\end{figure}

The array used for experimentation is composed
of eight microphones arranged on the summits of a rectangular prism.
The array is mounted on an ActivMedia Pioneer 2 robot, as shown in
Figure \ref{cap:array-pioneer2}. However, due to processor and space
limitations (the acquisition is performed using an 8-channel PCI soundcard
that cannot be installed on the robot), the signal acquisition and
processing is performed on a desktop computer (Athlon XP 2000+). The
algorithm currently requires 30\% CPU to work in real-time, but this
amount could be reduced by lowering the grid resolution or by using
approximations in computing the source probabilities. It is worth
mentioning that the CPU time does not increase with the number of
sources.

For all results presented in this paper, we used real multi-channel
recordings in a noisy environment with moderate reverberation. The
system is tested under different conditions. First, we measure the
maximum distance at which the system is able to detect different sound
sources. During the test, the sound source is produced 50 times with
the robot placed in different positions. The source detection rates
(number of detections/number of occurrences) are shown in Table \ref{cap:Detection-rate}.
We note that the system is able to reliably detect sources at distances
up to 5 meters. Also, while the system is able to detect bursts of
white noise reliably at great distance, it is mostly unable to detect
pure tones. This behavior is explained by the fact that sinusoids
occupy only a very small region of the spectrum and thus have a very
small contribution to the cross-correlations, even with the proposed
weighting. It must be noted that tones tend to be difficult to localize
even for the human auditory system.

\begin{table}[th]

\caption{Detection rate as a function of distance for different sounds\label{cap:Detection-rate}}

\center{\begin{tabular}{|c|c|c|c|}
\hline 
Sound source&
3 m&
5 m&
7 m\tabularnewline
\hline
\hline 
Hands clapping&
92\%&
94\%&
84\%\tabularnewline
\hline 
Speech ({}``test'')&
100\%&
90\%&
42\%\tabularnewline
\hline 
Noise burst (250 ms)&
100\%&
100\%&
100\%\tabularnewline
\hline
\end{tabular}}
\end{table}

\begin{figure*}[th]
\includegraphics[%
  width=0.28\paperwidth,
  keepaspectratio]{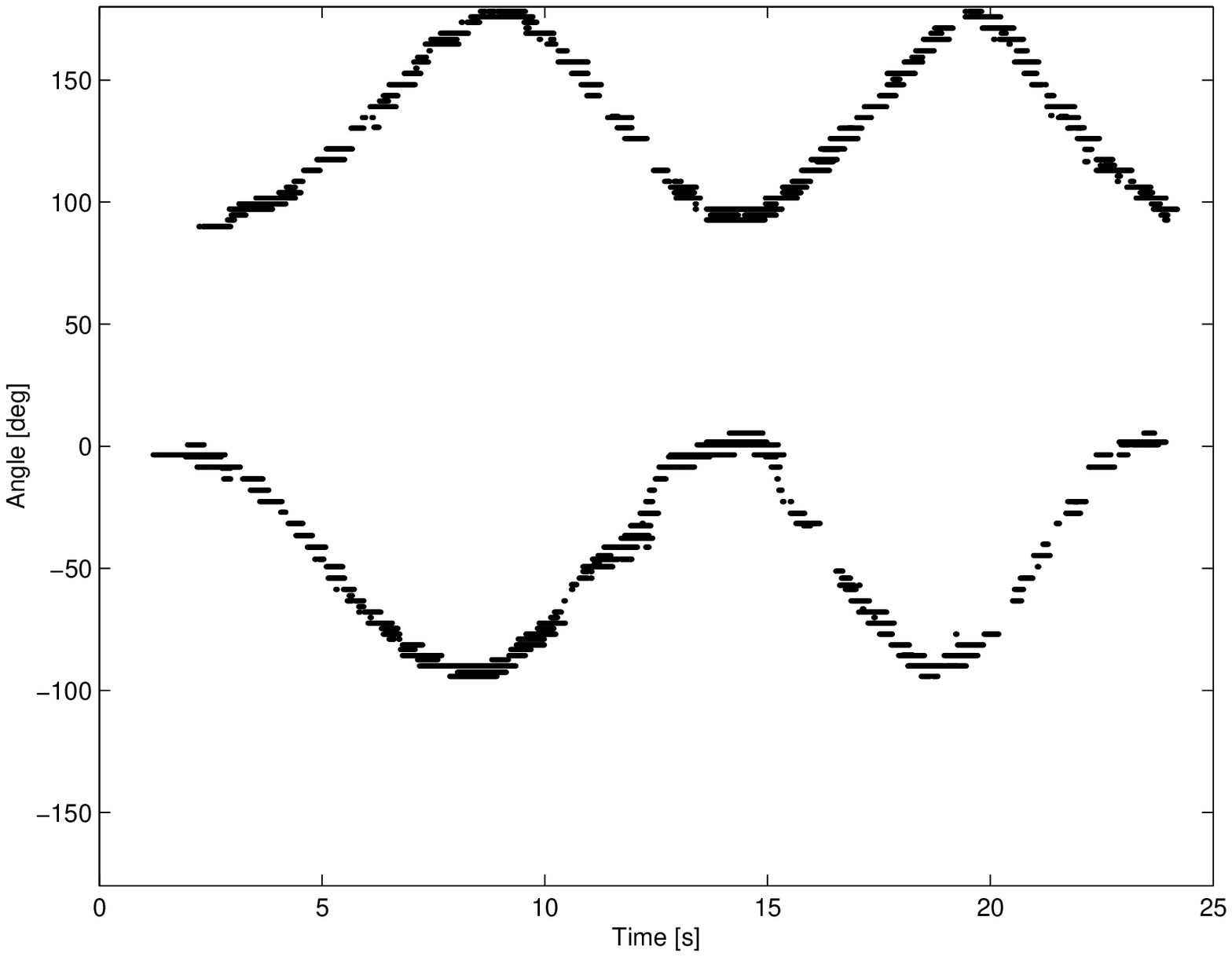}\includegraphics[%
  width=0.28\paperwidth,
  keepaspectratio]{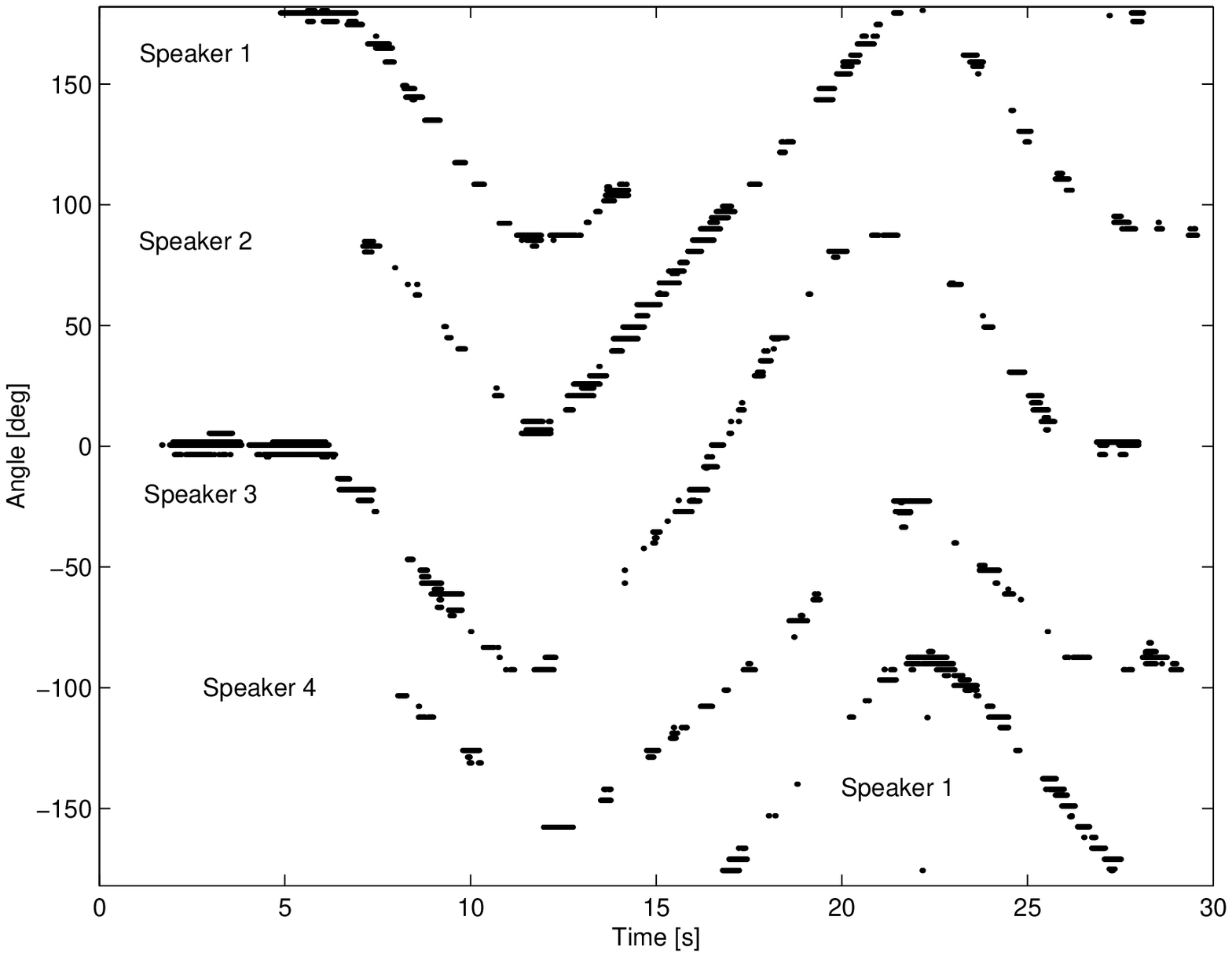}\includegraphics[%
  width=0.28\paperwidth,
  keepaspectratio]{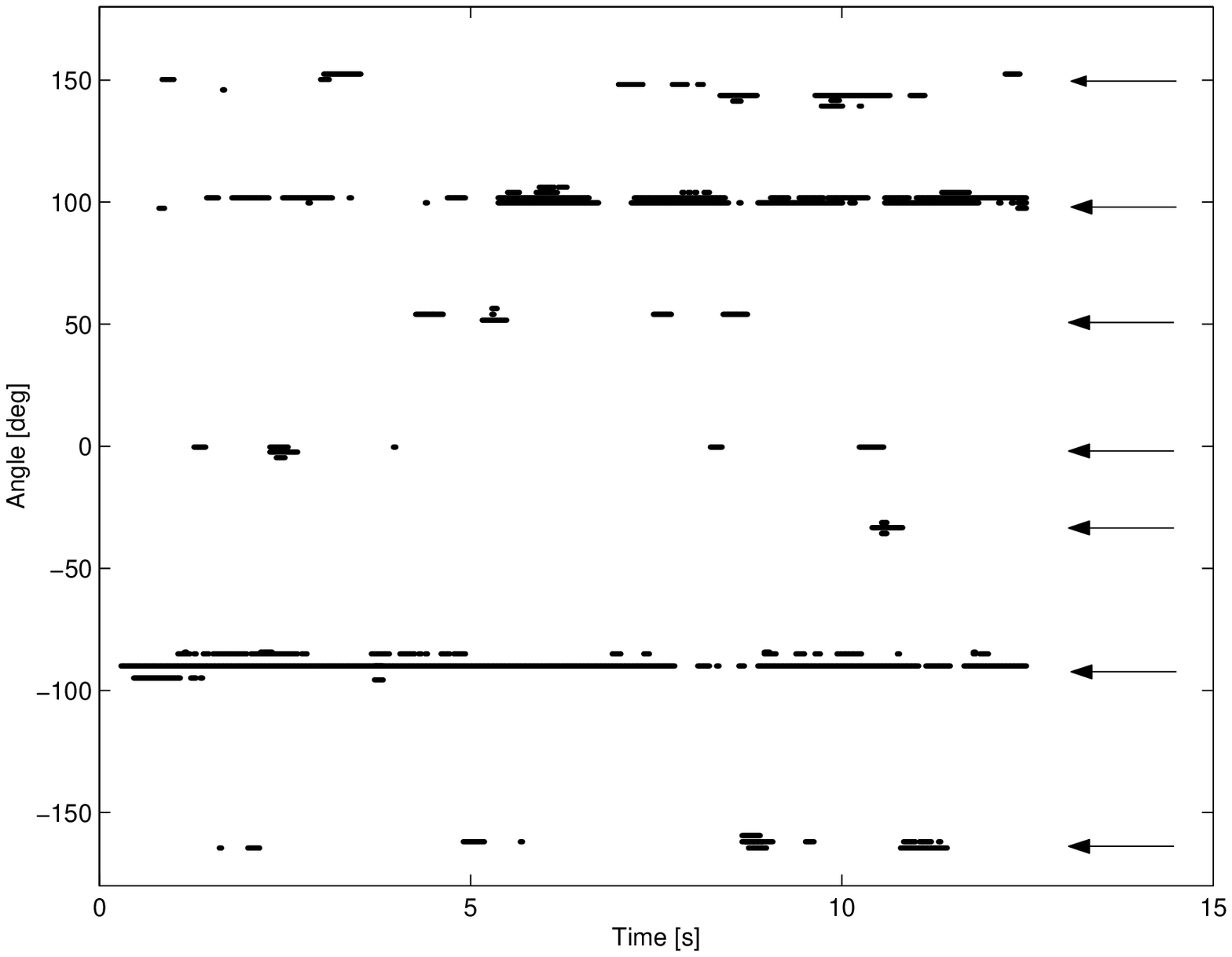}

\caption{Tracking of speech at a distance of 1-2 meters. a) 2 moving speakers
b) 4 moving speakers c) 7 stationary speakers (positions denoted by
arrows).\label{cap:Tracking-of-different-sources}}
\end{figure*}

The second task for which the system is evaluated is speaker tracking.
In this experiment, several people talk to the robot simultaneously
and in two of the three cases presented, the speakers are moving while
they talk. In Figure \ref{cap:Tracking-of-different-sources}, we
plot the regions where the probability of source presence is at least
0.6. Only azimuth is shown, since the sources are all located in the
same elevation range. From the Figure, it can be observed that the
system has no difficulty tracking up to 4 moving speakers. With 7
speakers, the system becomes unable to detect all speakers simultaneously,
but nonetheless succeeds in localizing them all at over a period of
time.

\begin{figure*}[th]
\includegraphics[%
  width=0.42\paperwidth,
  height=0.50\columnwidth]{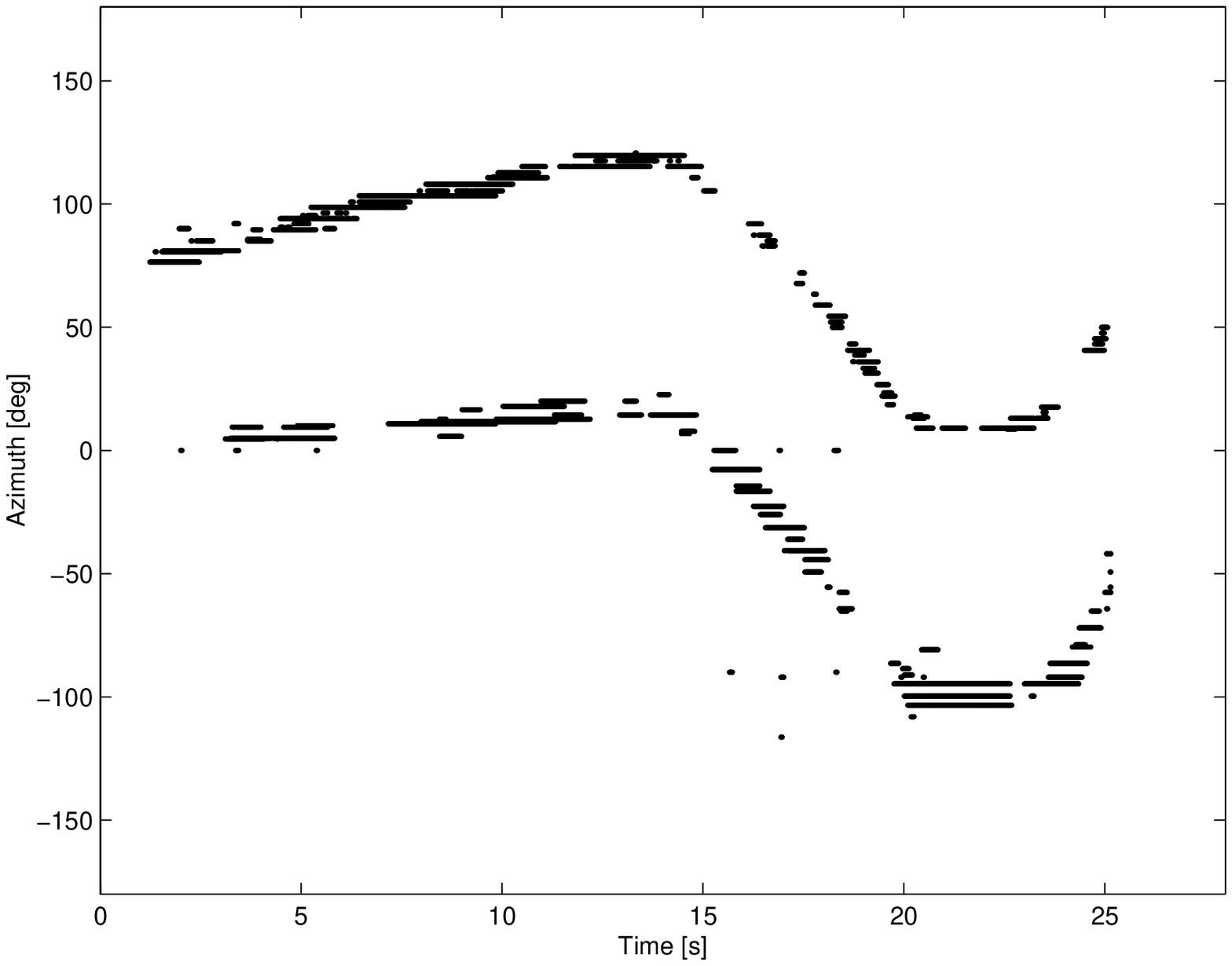}\includegraphics[%
  width=0.42\paperwidth,
  height=0.50\columnwidth]{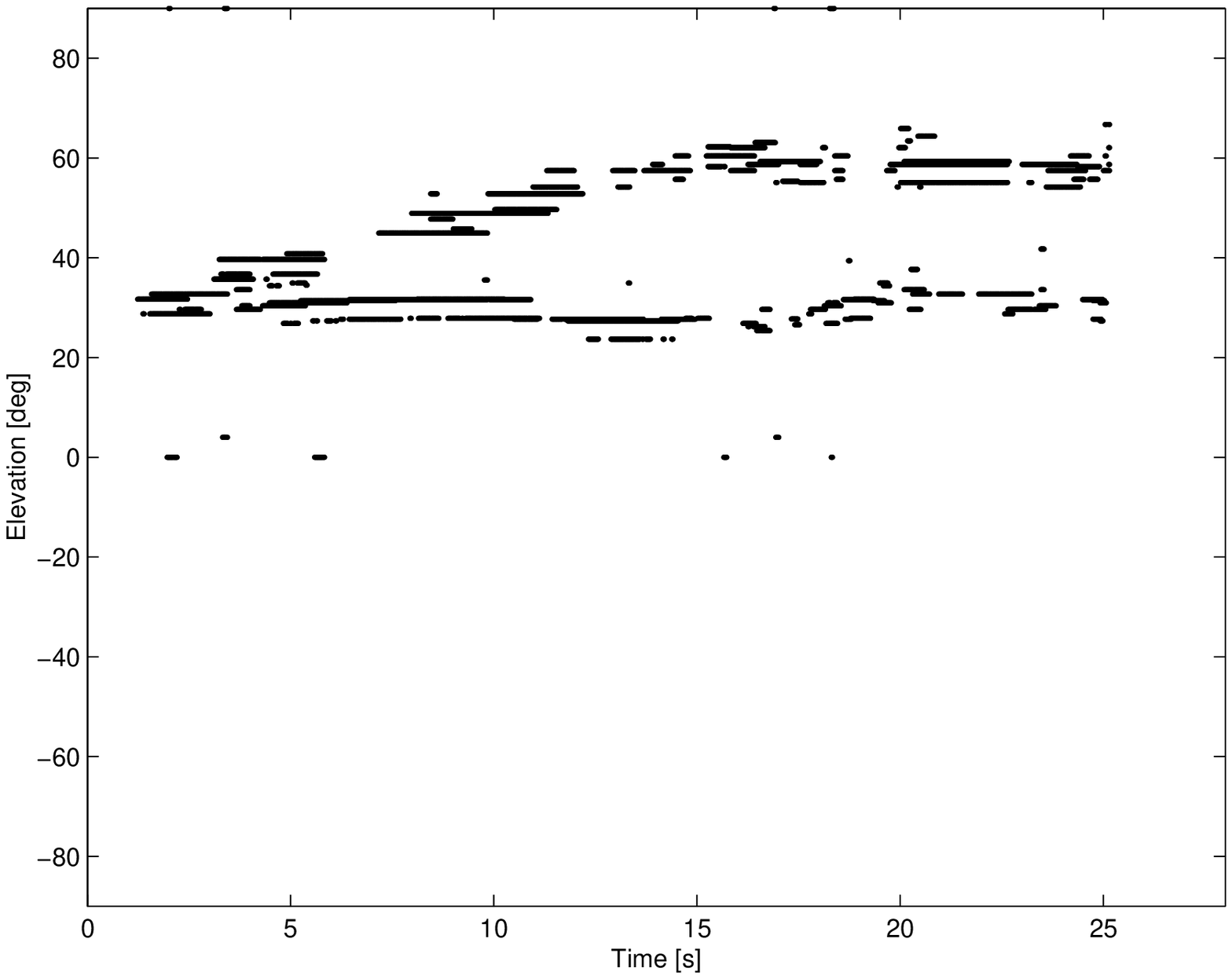}

\caption{Two stationary speakers with robot moving and rotating a) Azimuth
of sources b) Elevation of sources\label{cap:Two-stationary-speakers}}
\end{figure*}

A third test is performed with two stationary speakers, and a moving
robot. Figure \ref{cap:Two-stationary-speakers} shows how the robot
localizes the speakers as it moves. This demonstrates that the system
is able to function despite the noise caused by its motors. The two
sources that are sometimes detected at $0^{\circ}$ and $90^{\circ}$
elevation are respectively a computer fan located at 1.5 meter and
a ceiling ventilation trap.

A last experiment was conducted in which we verified that the system
still works when the microphone array is not completely open. Even
when some sides of the array are filled and some microphones no longer
have a line of sight with the source, the system's reliability is
not significantly affected.

\section{Conclusion}

Using an array of 8 microphones, we have implemented a system that
is able to reliably localize sounds up to five meters away, even in
the presence of noise. It is also possible to detect and track simultaneous
and moving sound sources. Moreover, our system is adapted to both
short-duration sounds like handclaps and longer duration sounds like
speech. 

In the proposed system, localization is performed in two steps. The
first step consists of a beamformer that is steered in all possible
directions, trying to maximize output power. The second step uses
Bayesian probability combinations to enhance the steered beamformer
results, removing most of the false detections while maintaining a
good detection rate.

In its current form, the localization system is very sensitive and
is sometimes able to detect weak sounds like computer fans located
within 2-3 meters. While this may in some cases be desirable, it may
be desirable in the future to design an algorithm capable of ranking
sound sources in terms of potential interest to the robot.

\section*{Acknowledgment}

Fran\c{c}ois Michaud holds the Canada Research Chair (CRC) in Mobile
Robotics and Autonomous Intelligent Systems. This research is supported
financially by the CRC Program, the Natural Sciences and Engineering
Research Council of Canada (NSERC) and the Canadian Foundation for
Innovation (CFI). Special thanks to Dominic L\'{e}tourneau, Serge
Caron, Nicolas B\'{e}gin, Mathieu Lemay, Pierre Lepage and Nathan
Sharfi for their help in this work. 

\bibliographystyle{/home/jm/phd/jmvalin/doc/icra2004/IEEEtran}
\bibliography{iros,BiblioAudible}

\end{document}